\documentclass{article}
\pdfoutput=1
\usepackage{microtype}
\usepackage{graphicx}
\usepackage{subfigure}
\usepackage{booktabs} 
\usepackage{amsmath}
\usepackage{amsthm}
\usepackage{amsfonts}

\theoremstyle{definition}

\usepackage{diagbox}
\usepackage{enumitem}

\newcommand{\squishlist}{
   \begin{list}{$\bullet$}
    { \setlength{\itemsep}{0pt}      \setlength{\parsep}{0pt}
      \setlength{\topsep}{0pt}       \setlength{\partopsep}{0pt}
      \setlength{\listparindent}{-2pt}
      \setlength{\itemindent}{-5pt}
      \setlength{\leftmargin}{1em} \setlength{\labelwidth}{0em}
      \setlength{\labelsep}{0.5em} } }
\newcommand{\squishend}{
    \end{list}  }

\newcommand{\todo}[1]{\textcolor{black}{#1}}

\usepackage{hyperref}

\usepackage[accepted]{mlsys2021}

\mlsystitlerunning{Understanding GNN Computational Graph: A Coordinated Computation, IO, and Memory Perspective}

\newcommand{\Scatter}{\texttt{Scatter}}
\newcommand{\Gather}{\texttt{Gather}}
\newcommand{\MLPEdge}{\texttt{ApplyEdge}}
\newcommand{\MLPVertex}{\texttt{ApplyVertex}}
\newcommand{\FusedVtoV}{\texttt{Aggregate}}
\newcommand{\FusedEtoE}{\texttt{ReduceScatter}}
\newcommand{\MLPs}{\texttt{Apply-}}
\newcommand{\LightApply}{lightweight {\MLPs}}
\newcommand{\ExpensiveApply}{expensive {\MLPs}}
\newcommand{\Gchar}{\mathcal{G}}
\newcommand{\Vchar}{\mathcal{V}}
\newcommand{\Echar}{\mathcal{E}}

\newcommand{\pleasecheck}[1]{\textcolor{black} {#1} \textcolor{black}}
\newcommand{\grad}[1]{\text{grad}#1}
\newcommand{\graphirr}{\pleasecheck{graph-irrelevant}}

\begin{document}

\twocolumn[
\mlsystitle{Understanding GNN Computational Graph: A Coordinated Computation, IO, and Memory Perspective}



\mlsyssetsymbol{equal}{*}

\begin{mlsysauthorlist}
\mlsysauthor{Hengrui Zhang}{equal,thu}
\mlsysauthor{Zhongming Yu}{equal,thu}
\mlsysauthor{Guohao Dai}{thu}
\mlsysauthor{Guyue Huang}{ucsb}
\mlsysauthor{Yufei Ding}{ucsb}
\mlsysauthor{Yuan Xie}{ucsb}
\mlsysauthor{Yu Wang}{thu}

\end{mlsysauthorlist}

\mlsysaffiliation{thu}{Tsinghua University}
\mlsysaffiliation{ucsb}{University of California, Santa Barbara}

\mlsyscorrespondingauthor{Guohao Dai}{daiguohao@mail.tsinghua.edu.cn}
\mlsyscorrespondingauthor{Hengrui Zhang}{hengrui-18@mails.tsinghua.edu.cn}

\mlsyskeywords{Machine Learning, MLSys}

\vskip 0.3in

\begin{abstract}

Graph Neural Networks (GNNs) have been widely used in various domains, and GNNs with sophisticated computational graph lead to higher latency and larger memory consumption. Optimizing the GNN computational graphs suffers from:
\textbf{(1) Redundant neural operator computation.} The same data are propagated through the graph structure to perform the same neural operation multiple times in GNNs, leading to redundant computation which accounts for 92.4\% of total operators.
\textbf{(2) Inconsistent thread mapping.} Efficient thread mapping schemes for vertex-centric and edge-centric operators are different. This inconsistency prohibits operator fusion to reduce memory IO.
\textbf{(3) Excessive intermediate data.} For GNN training which is usually performed concurrently with inference, intermediate data must be stored for the backward pass, consuming 91.9\% of total memory requirement.


To tackle these challenges, we propose following designs to optimize the GNN computational graph from a novel coordinated computation, IO, and memory perspective:
\textbf{(1) Propagation-postponed operator reorganization.} We reorganize operators to perform neural operations before the propagation, thus the redundant computation is eliminated.
\textbf{(2) Unified thread mapping for fusion.} We propose a unified thread mapping scheme for both vertex- and edge-centric operators to enable fusion and reduce IO.
\textbf{(3) Intermediate data recomputation.} Intermediate data are recomputed during the backward pass to reduce the total memory consumption.
Extensive experimental results on three typical GNN models show that, \todo{we achieve up to 2.75$\times$ end-to-end speedup, 6.89$\times$ less memory IO, and 7.73$\times$ less memory consumption over state-of-the-art frameworks.}


\end{abstract}
]



\printAffiliationsAndNotice{\mlsysEqualContribution} 

\section{Introduction}

Graph Neural Networks (GNNs) explore features of vertices and edges using neural operators and relationships through the graph structure. GNNs have shown great potentials in various domains, including Recommendation Systems~\cite{pinsage, wang2019knowledge}, Computer Vision~\cite{yan2018spatial, qi2018learning}, Natural Language Processing~\cite{nguyen2018graph, yao2018graph},  \emph{et al}~\cite{gcn, graphsage}.

With the fast development of GNNs, GNN models have evolved into more diversity and complexity in the computational graph, putting forward expensive requirements on both computation and memory resources. For example, training a GNN-based recommendation model consumes 16 GPUs (384 GB memory in total) using days of time~\cite{pinsage}. Improving the performance of GNNs with less resources suffers from: (1) From the computation perspective, GNN models perform neural operators through the graph structure, meaning that the same data of a vertex may be propagated to different edges. Thus, the same operation applied on these edges are executed multiple times for the same vertex data after propagation, leading to \textbf{redundant computation} in GNNs. We measure that such redundant computation account for 92.4\% of total operators in an EdgeConv model~\cite{edgeconv}, with the detailed setup in Section~\ref{sec:exp}. (2) From the IO perspective, current systems involve writing/reading graph-sized feature data between two graph operators. Operators performed on vertices and edges are usually with \textbf{inconsistent thread mapping} schemes, which hinder applying fusion for these operators to reduce IO. (3) From the memory perspective, GNN models usually perform concurrent training and inference passes. Thus, \textbf{excessive intermediate data} produced during executing fused operators must still be stored for backward, leading to large memory space requirement. We measure in a Graph Attention Network (GAT)~\cite{gat} model that the intermediate data consume 91.9\% of total memory.

To tackle these challenges and accelerate GNN computation with less memory consumption, we need a systematic GNN computational graph optimization framework which considers computation, IO, and memory. DGL~\cite{dgl} provides two high-level operators, gSpMM and gSDDMM, to express various GNN models, while such an abstraction fails to explore the redundant computation hidden in neural operators performed through the graph structure. FuseGNN~\cite{fusegnn} fuses edge operators to accelerate GNN computation, but it lacks the technique to fuse a vertex-centric operator with an edge-centric one. Huang \textit{et al.,}~\cite{ppopp2021} also proposes fusion technique for GNNs, while it cannot handle GNN training because the intermediate data are missing. 

All previous studies fail to comprehensively consider the computation, IO, and memory perspectives for both GNN training and inference. Thus, we put forward a systematic framework to accelerate GNN computation and reduce memory consumption on GPUs with following contributions: 

\squishlist

\item \textbf{Propagation-postponed operator reorganization.} Since the redundant computation is caused by performing neural operators through the graph structure, we reorganize operator to perform neural operations before the propagation with an average of \todo{1.68}$\times$ speedup. 

\item \textbf{Unified thread mapping for fusion.} Since different thread mapping scheme prohibits fusing vertex-centric and edge-centric operator and further reducing IO, we propose a unified thread mapping scheme for both types of operators and save up to \todo{5.45$\times$} memory IO.

\item \textbf{Intermediate data recomputation.} Since the intermediate data consume the majority of memory but are only stored for the backward pass, we introduce a recomputation mechanism to reproduce intermediate data just before they are needed in backward use. We save the memory consumption by up to \todo{2.21$\times$}.

\squishend

\begin{figure}[!tp]
    \centering
    \includegraphics[width=0.46\textwidth]{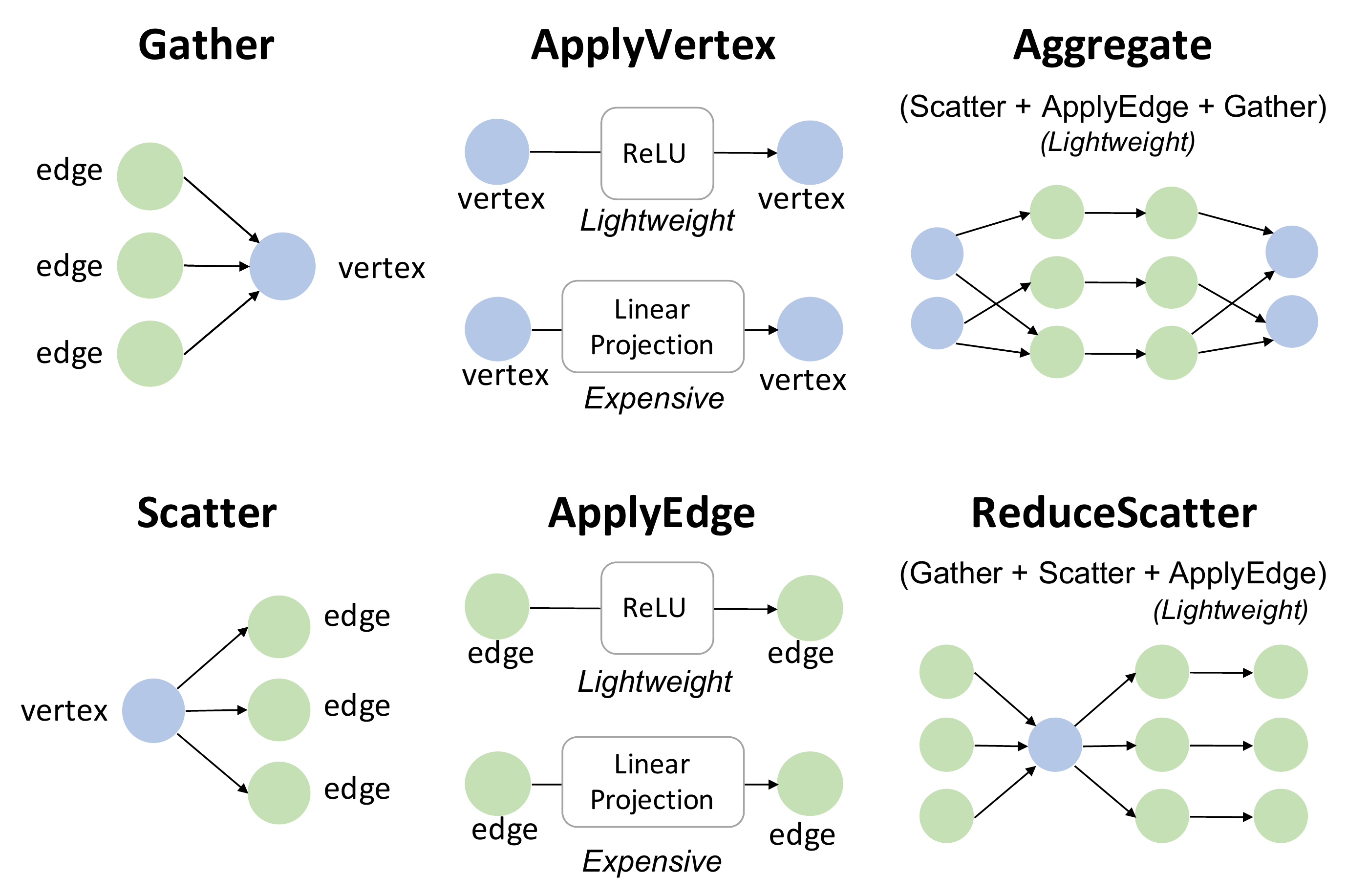}
    \vspace{-15pt}
    \caption{Operators in Graph Neural Networks (GNNs).}
    \vspace{-20pt}
    \label{fig:op}
\end{figure}

We implement three popular GNN models with the techniques above, achieving up to \todo{2.75}$\times$ speedup, \todo{6.89}$\times$ less IO, and \todo{7.73}$\times$ less memory consumption. We even enable running  large-scale GNN models with an NVIDIA RTX 2080 GPU (8 GB), which requires the newest NVIDIA RTX 3090 GPU (24 GB), with a comparable latency. 

The following of this paper is organized as follows. Section~\ref{sec:bg} introduces preliminaries of GNNs. Section~\ref{sec:overview} introduces an overview of our optimization recipe. Our three techniques, propagation-postponed operator reorganization, unified thread mapping for fusion, and intermediate data recomputation are detailed in Section~\ref{sec:reorg}, \ref{sec:fusion}, and \ref{sec:recompute}, respectively. Section~\ref{sec:exp} presents evaluation results. Section~\ref{sec:relatework} elaborates related work. Section~\ref{sec:conclusion} concludes the paper.



\section{Preliminaries}\label{sec:bg}


\subsection{GNN Operators}
On a graph $\Gchar=(\Vchar, \Echar)$ with the set of vertices $\Vchar$ and edges $\Echar$, a GNN layer is composed of the following operators:

\vspace{-10pt}
{
\small
\begin{equation*} \begin{aligned}
    m_e &= \Scatter (h_v, h_u), (u,e,v) \in \Echar, \\
    m^{new}_e &= \MLPEdge (m_e[, m'_e, \cdots]),\\
    h_v &= \Gather (\{m_e: (u,e,v) \in \Echar \}),\\
    h^{new}_v &= \MLPVertex (h_v[, h'_v, \cdots]).
\end{aligned} \end{equation*}
}
\vspace{-10pt}

In the above equations, $v, u$ are vertex indices and $e$ is an edge index. $h_v$ refers to feature attached to vertex $v$, and $m_e$ attached to edge $e$. 

Figure~\ref{fig:op} visualizes the definitions of operators. {\Gather} is a reduction operation that generates the feature of a vertex from features of edges connecting to it. {\Scatter} generates the feature of an edge from features of vertices that the edge connects to. {\MLPEdge} and {\MLPVertex} are \graphirr operators that transform the features of each edge and vertex, respectively. We further categorize {\MLPs} operators based on their computation cost: element-wise operations are considered as {\LightApply}, while computation-intensive operations like linear projection are considered {\ExpensiveApply}.

The four operators above are \pleasecheck{comprehensive enough} to express any GNN model, but there are some widely-used combinations of operators, which current GNN systems also provide dedicated optimizations to. We name two most common combinations: {\FusedVtoV} and {\FusedEtoE}, as defined below. We add them to our operator abstraction operators for the convenience of expressing models.

\vspace{-15pt}
{
\small
\begin{equation*} \begin{aligned}
    h^{new}_v &= \FusedVtoV(\{(h_u, m_e): (u,e,v) \in \Echar\}, h_v) \\
    &= \Gather(\{\MLPEdge(\Scatter(h_v, h_u), m_e)\}) \\
    m^{new}_e &= \FusedEtoE(\{m_{e'}: (u\in N(v), e', v) \in \Echar \}, h_u)\\
    &= \MLPEdge(\Scatter(\Gather(\{m_{e'}\}),h_u), m_e),\\
    &(u,e,v) \in \Echar
\end{aligned} \end{equation*}
}
\vspace{-15pt}

{\FusedVtoV} generates a new vertex feature by reducing features from its neighbor vertices and edges. {\FusedEtoE} generates a new edge feature by reducing and scattering among the group of edges that connect to the same vertex, a typical example being the edge-softmax operation in the Graph Attention Network (GAT). Current GNN systems widely support fused {\FusedVtoV} and {\FusedEtoE} implementations when {\MLPEdge} is lightweight \cite{gespmm,dgl}.

\textbf{Compared with related work, our operator abstraction is both comprehensive and optimization-friendly.} In terms of comprehensiveness, the Aggregation-Combination abstraction in previous work ~\cite{yan2020hygcn, gnnadvisor}, equivalent to our {\FusedVtoV} and {\MLPVertex}, does not cover {\MLPEdge}. Therefore, the Aggregation-Combination can only express GNN models without applying neural networks to edge features, such as the vanilla GCN~\cite{gcn} or GraphSAGE~\cite{graphsage}. Our proposed operator abstraction, in constrast, can construct whatever Aggregation-Combination constructs, and also Graph Attention Network (GAT)~\cite{gat}, EdgeConv~\cite{edgeconv}, and other models with arbitraty message-passing procedure. Figure~\ref{fig:overall}(a) shows how to construct GAT using our operator abstraction, and the construction of more GNN models are elaborated in Appendix to demonstrate its comprehensiveness. In terms of optimization convenience, the abstraction in DGL~\cite{dgl}, gSDDMM and gSpMM, can be lowered to any operator-combination that outputs edge and vertex features. respectively. Such general abstraction hinders chances of local, global or adaptive optimizations, e.g. optimizing only {\Gather} part, or fusing the last {\Scatter} in gSDDMM with first {\Gather} in gSpMM. DGL leverages a limited set of built-in operators to tackle optimization challenges in such a general abstraction. On the contrary, this paper uses a fine-grained operator abstraction to express GNN models for generality, and leverage inter-operator optimizations to systematically improve performance. 

\begin{figure}[!tp]
    \centering
    \includegraphics[width=0.45\textwidth]{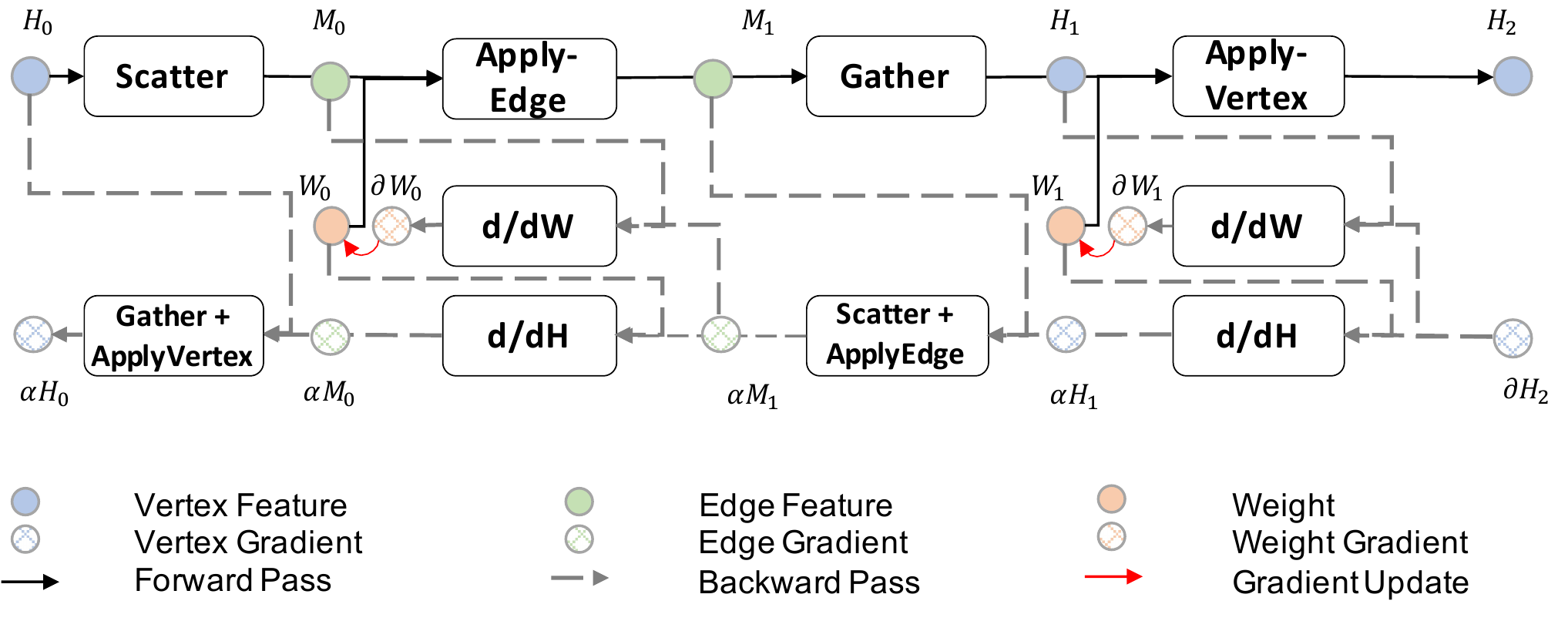}
    \vspace{-12pt}
    \caption{Dataflow in GNN training, showing both the forward pass (top) and backward pass (bottom). All intermediate features are used for backward, which need to be stashed in the memory.}
    \vspace{-20pt}
    \label{fig:backprop}
\end{figure}

\subsection{Back-Propagation in GNN}
The back-propagation algorithm is applied to train GNN models. One can prove that the backward pass of above set of operators still fall into this set. We list the key conclusions below, while the detailed proof is elaborated in Appendix. 

\squishlist
    \item The backward pass of {\Gather} ({\Scatter}) is {\Scatter} and {\MLPVertex} ({\Gather} and {\MLPEdge}).
    \item The backward pass of {\MLPEdge} ({\MLPVertex}) is two {\MLPEdge} ({\MLPVertex}) operations.
\squishend

\begin{figure*}[!tp]
    \centering
    \includegraphics[width=0.9\textwidth]{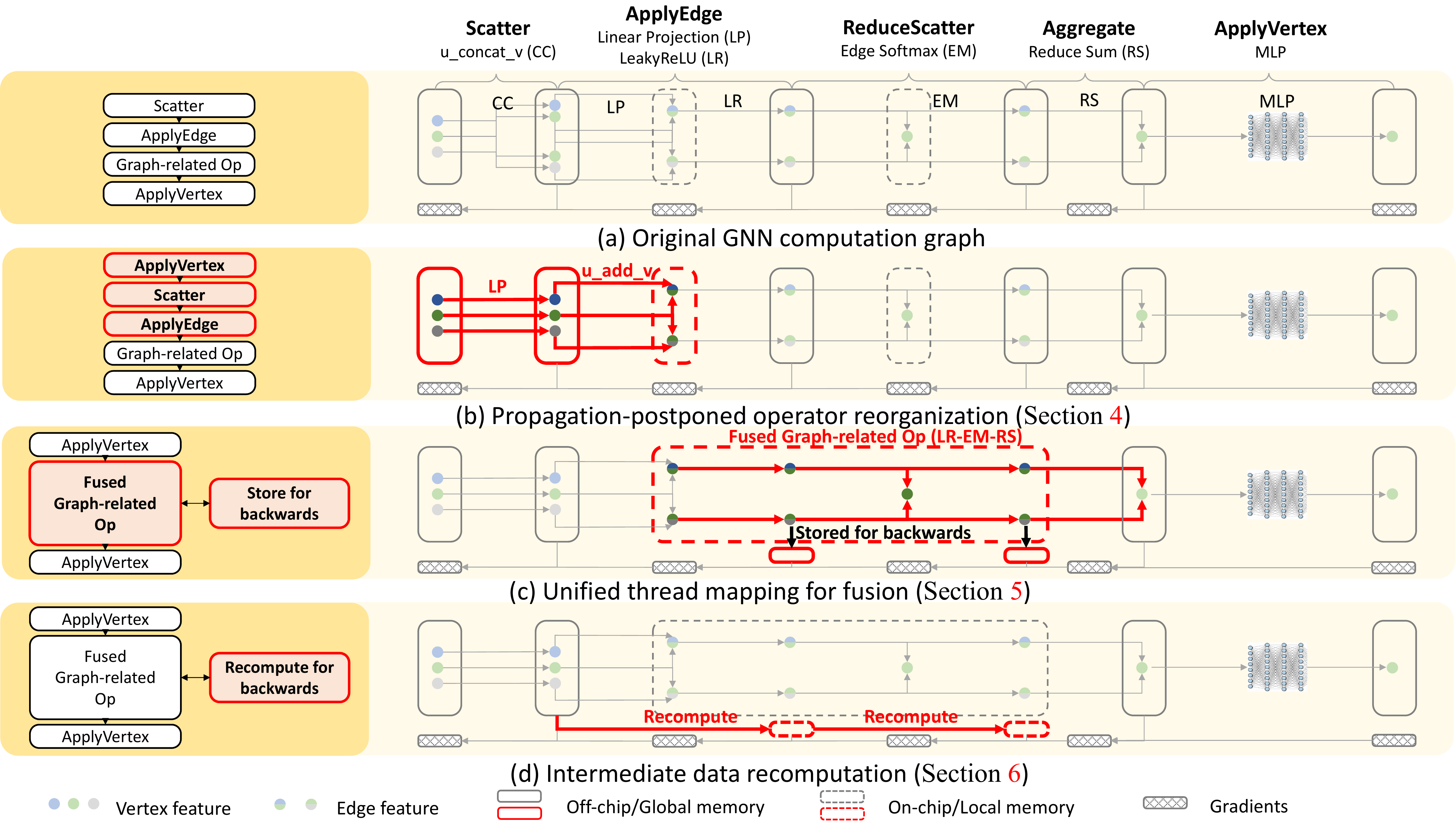}
    \vspace{-10pt}
    \caption{Design Overview. The left part shows the high-level abstraction of a typical GNN computation flow, and the right part shows the example of a GAT~\cite{gat} model when applying techniques proposed in this paper. (a) The original computation graph. (b) After applying operator reorganization, the linear projection operator is preposed and applied on vertices to reduce computation. (c) After applying operator fusion with the unified thread mapping scheme, operators are fused while the intermediate are still stored for the back propagation phase. (d) After applying recomputation, intermediate data are required to be stored.}
    \vspace{-15pt}
    \label{fig:overall}
\end{figure*}

The backward pass of {\FusedVtoV} ({\FusedEtoE}) can be analyzed by decomposing them into the four fine-grained operators. In summary, we can express both forward and backward pass of GNN using the same operator abstraction. A dataflow showing both passed and the four basic operators are in Figure~\ref{fig:backprop}. Figure~\ref{fig:backprop} also show that the intermediate features are needed for computing gradients in the backward pass of the same layer. Therefore, all intermediate features need to be stashed in the memory for backward. During the forward pass, all intermediate features must be saved and later used to calculate parameter gradients in the backward pass. Take the GAT example again, Figure~\ref{fig:overall}(a) marks all the  feature tensors that are stashed and where they are used in the backward pass. 
State-of-the-art GNN systems follow the general strategy of saving outputs of all operators in the model, and only provide fused implementations for some common operator-combinations to avoid saving an $O(|\mathcal{E}|)$ intermediate tensor (e.g., DGL's built-in edge-softmax for GAT), a general approach for reducing memory consumption in training is lacked. 

\section{Design Overview}\label{sec:overview}
This paper proposes a systematic approach to optimize the GNN at inter-op level. In this section, we provide an overview of our designs by walking-through them on the model architecture of Graph Attention Network (GAT)~\cite{gat}. As Figure~\ref{fig:overall}(a) shows, a GAT layer is composed of {\Scatter}, {\MLPEdge}, {\FusedEtoE},  {\FusedVtoV}, ended by an {\MLPVertex}. We tackle the aforementioned three design challenges with methods below:

\textbf{Eliminating redundant computation through propagation postponed reorganization}. Recall that the redundancy is caused by performing {\ExpensiveApply} (e.g. linear transformations) many times on the features propagated from the same source. We propose to reorder the operator sequence: to first apply  {\ExpensiveApply} on the vertex features, and next propagate the result of the operation. For example, we show that in GAT~\cite{gat} models, the {\Scatter}-{\MLPEdge} operator sequence can both be substituted by linear-projection on vertex features and {\Scatter}-ing the result, as shown in Figure~\ref{fig:overall}(b).

 \textbf{Reducing IO through completely fusing graph-related kernels.} We propose to fuse a sequence of operators as long as they are graph-related kernels or {\LightApply}. We choose not to fuse {\ExpensiveApply} like linear transformations because they can often be tackled with primitives in highly-optimized libraries, e.g. cuBLAS or cuDNN. The challenge here is that vertex-centric and edge-centric operators, i.e. operators that produce vertex- or edge-features, apply vertex-balanced and edge-balanced thread mapping in current GNN systems, respectively. The unmatched thread mapping schemes prohibit reusing intermediate data locally and force dumping data to the DRAM. With novel kernel designs, we show vertex- and edge-balanced thread mapping can both be applied no matter the operator produces vertex- or edge-features. This allows us to choose a unified thread mapping for a sequence of graph-related kernels to fuse them. As shown in Figure~\ref{fig:overall}(c), this step fuses operators like {\Scatter}, {\FusedEtoE}, {\FusedVtoV} into one single kernel and greatly reduces IO.
 
\textbf{Avoiding saving intermediate data for backward pass through recomputation.} Recall that GNN training requires saving all intermediate features in the forward pass for computing gradients, leading to excessive memory consumption. We borrow the idea of gradient checkpointing in DNN training to tackle this challenge. We selectively save the intermediate features in the forward pass (checkpoints), and recompute the unsaved features just before they are needed in the backward pass,  as shown in Figure~\ref{fig:overall}(d). The non-checkpoint features originally requires a memory size of $O(d \times |\mathcal{E}|)$, where $d$ stands for the number of GNN layers and $|\mathcal{E}|$ is the number of edges. With recomputation and the aforementioned kernel-fusion technique, we can eliminate this $O(d \times |\mathcal{E}|)$. To maximize the benefit of memory savings and minimize the cost of recomputation, we choose to recompute edge rather than vertex features. 



\section{Reducing Computation: Propagation-postponed Operator Reorganization \label{sec:reorg}
}
\textbf{Motivation.}
Many GNN models perform {\Scatter} followed by a computation-intensive neural network (NN) as {\MLPEdge}. The same vertex feature is propagated to all its adjacent edges, and this duplication causes repeated NN computation in the {\MLPEdge} step.

\textbf{Challenge.} We describe above the intuition why propagation + NN operator causes redundant computation, but we lack a general formulation to identify and eliminate such redundancy. In particular, {\Scatter} involves both redundant computation and \pleasecheck{per-edge unique} computation: the redundant part is because multiple edges connected to the same vertex share identical vertex feature as input, and the unique part is because each edge combines features from a unique pair of two vertices. Separating the two parts and reducing the redundant part require a careful surgery on the original computational graph. 

\begin{figure}[!tp]
    \centering
    \includegraphics[width=0.48\textwidth]{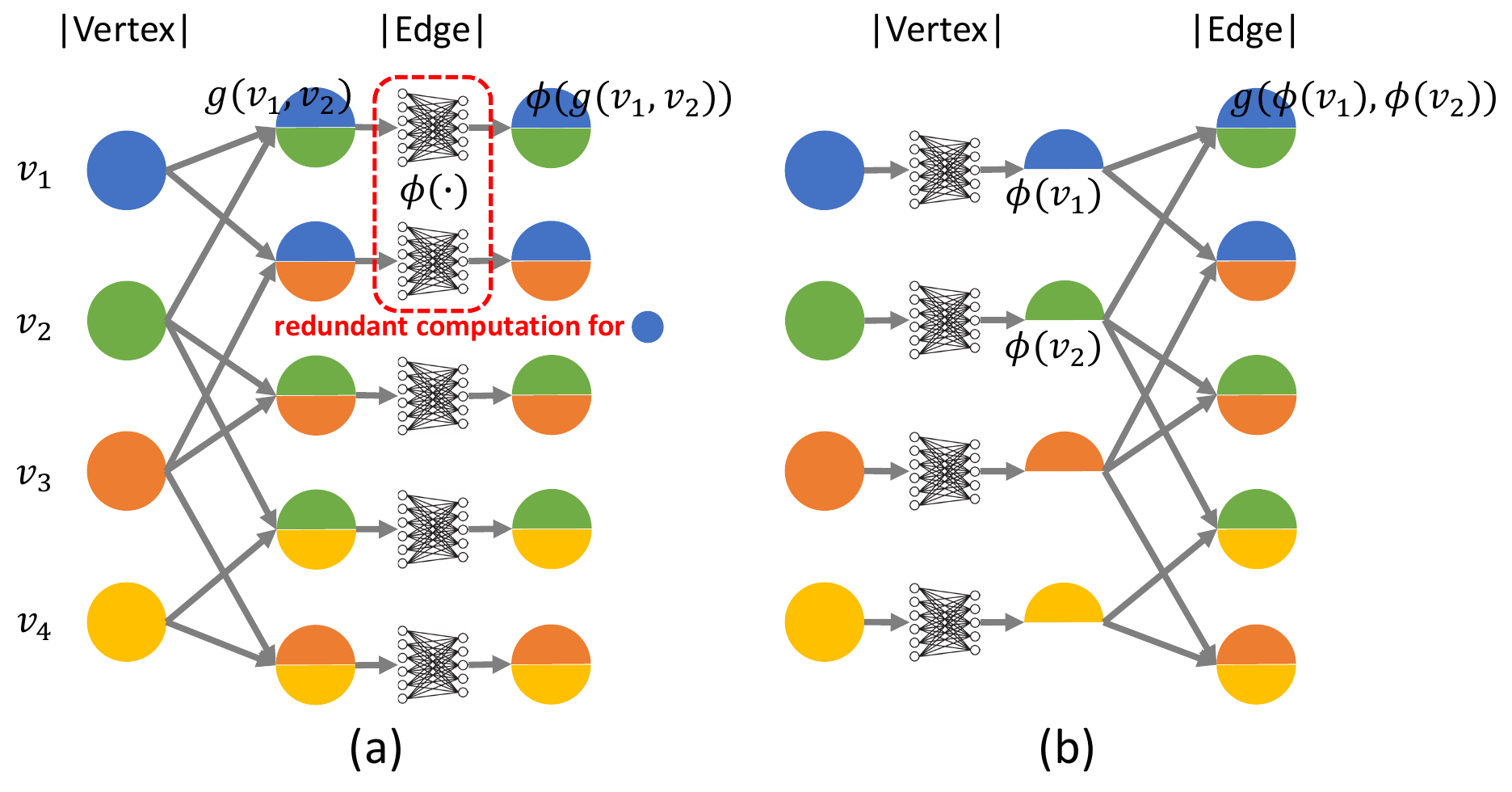}
    \vspace{-25pt}
    \caption{Diagram of the propagation-postponed operator reorganization. (a) Redundant neural operator computation on a same vertex. (b) Operation reorganization to postpone graph operators and eliminate redundant computation.}
    \vspace{-15pt}
    \label{fig:reorg}
\end{figure}

\textbf{Insight.} Our key insight is that the root of this possible computation redundancy is {performing repeated neural computation on features scattered from the same source}. Take figure \ref{fig:reorg}(a) as an example. Figure \ref{fig:reorg}(a) shows the computation and data flow for a part of one EdgeConv layer with one {\Scatter} operator followed by an {\MLPEdge} operator. Features on vertices are first scattered to edges with function $g(u,v)=u-v$, after that a linear-projection function $\phi(\cdot)$ is applied. Vertex features are scattered and applied $\phi(\cdot)$ independently on different edges. Therefore we might apply $\phi(\cdot)$ to the same feature more than once, which causes possible redundancy in computation.

\textbf{Approach: identify redundancy.} Following our insight that the possible redundancy occurs in the {\Scatter}-{\MLPEdge} phase, we find a sufficient condition to identify this possible redundancy in computing: if the Scatter function $g$ and {\MLPEdge} function $\phi$ follows the \textbf{commutative law} and \textbf{distributive law}, there is redundancy in this {Scatter-ApplyEdge} procedure.
Take figure \ref{fig:reorg}(a) as an example. We first compute $g(v_1,v_2)$ and $g(v_1,v_3)$ during Scatter, then compute $\phi(g(v_1,v_2))$ and $\phi(g(v_1,v_3))$. Under the commutative law and distributive law, we obtain $\phi(g(v_1,v_2))=g(\phi(v_1),\phi(v_2))$ and $\phi(g(v_1,v_3))=g(\phi(v_1),\phi(v_3))$. Therefore, we actually compute $\phi(v_1)$ more than once. For the whole procedure, the computational expensive function $\phi(\cdot)$ is computed $|\mathcal{E}|$ times.

\textbf{Approach: eliminate redundancy}. 
We propose propagation postponed operator reorganization to eliminate this redundancy while keeping functional equivalence. The main idea is, as the redundancy is caused by edges that share the same source performing transformation to the same feature, if we postpone {\Scatter} and perform \texttt{ApplyFunction} first, we will only perform transformation to the same feature for only once. In  figure \ref{fig:reorg}(b), we first compute $\phi(v_1)$, $\phi(v_2)$ and $\phi(v_3)$, then scatter them to edges to compute $g(\phi(v_1),\phi(v_2))$ and $g(\phi(v_1),\phi(v_3))$, which actually change the execution order from {\Scatter}-{\MLPEdge} to {\MLPVertex}-{\Scatter}. For the whole procedure, function $g$ is still computed $|\mathcal{E}|$ times, but the computational expensive function $\phi(\cdot)$ is computed only $|\mathcal{V}|$ times. In most cases, $g$ is arithmetic operator and $\phi(\cdot)$ is linear operator, which means the distributive law and commutative law are met, and we can always eliminate this redundancy with operator propagation-postponed operator reorganization.

\textbf{Example. }
In GATConv, the {\Scatter}-{\MLPEdge} computes the attention score between two vertices by concatenating and applying a one-layer neural network mechanism, as in Equation \ref{eq:attn_score}:

\vspace{-10pt}
\begin{equation}
\label{eq:attn_score}
    {e}_{u\rightarrow v}=\operatorname{LeakyReLU}\left(\vec{a}^{T}\left[\vec{h}_{u} \| \vec{h}_{v}\right]\right)
 \vspace{-5pt}  
\end{equation}

where $\vec{h}_u,\vec{h}_v\in \mathbf{R}^{f}$ are the feature vector of the destination and source, and $\vec{a}\in\mathbf{R}^{2f}$ are weight parameters to learn.

As Figure \ref{fig:overall}(a) shows, a {\Scatter} operator $\operatorname{u\_concat\_v}$ is first applied to propagate features to edges and concatenate the feature vector of the source and destination into $\left[\vec{h}_{u} \| \vec{h}_{v}\right]$, followed by a $\operatorname{LP}$ (Linear Projection) and $\operatorname{LeakyReLU}$ to compute the final attention score. The computation cost is $2|\mathcal{E}|f$ for $\operatorname{u\_concat\_v}$, $4|\mathcal{E}|f$ for $\operatorname{LP}$ and $|\mathcal{E}|$ for $\operatorname{LeakyReLU}$ , with a total of $6|\mathcal{E}|f+|\mathcal{E}|$.

Although $\operatorname{concatenate}$ and the non-linear neural operation do not follow the commutative and distributive law, we find that the $\operatorname{LP}$ and $\operatorname{concatenate}$ can be seen as two $\operatorname{LP}$ followed by an $\operatorname{add}$: $\vec{a}^{T}\left[\vec{h}_{u} \| \vec{h}_{v}\right]=\left[\vec{a_l}^{T}\|\vec{a_r}^T\right]\left[\vec{h}_{u} \| \vec{h}_{v}\right]=\vec{a_l}^T\vec{h}_{u}+\vec{a_r}^T\vec{h}_{v}$ . Therefore, there is redundancy in computation if we don't perform operator reorganization, and we postpone Scatter and change the execution order from {\Scatter}-{\MLPEdge} into {\MLPVertex}-{\Scatter}. As shown in Figure~\ref{fig:overall}(b), we first apply $\operatorname{LP}$ to features on vertices, then scatter them to edges and perform $\operatorname{add}$, followed by a $\operatorname{LeakyReLU}$, which still need to be applied on edges. The total computation cost is reduced to $4|\mathcal{V}|f+2|\mathcal{E}|$.

\section{Reducing IO: Unified Thread Mapping for Fusion} \label{sec:fusion}

\textbf{Motivation.}
GNN systems suffer from excessive global memory writing/reading between production-consumption operators. Take the GAT model in Figure~\ref{fig:overall} as an example: the edge features produced by the {\MLPEdge} step needs to be written-out to the global memory, and read-in again by the next {\FusedEtoE} operator. The output of {\FusedEtoE} step is again stored and loaded by the succeeding {\FusedVtoV} kernel. Both procedures involve writing/reading a $O (| \mathcal{E} |)$-sized feature tensor. Kernel fusion is widely exploited to reduce the data movement. In fact, the edge-softmax in current systems are commonly implemented by a hand-optimized fused kernel to reduce IO. Our target is to apply kernel fusion to further eliminate the aforementioned two edge feature store/load, and completely fuse all graph-related operators ({\Scatter}, {\MLPEdge}, {\FusedEtoE}, {\FusedVtoV}). 

\begin{figure}[!tp]
    \centering
    \includegraphics[width=0.48\textwidth]{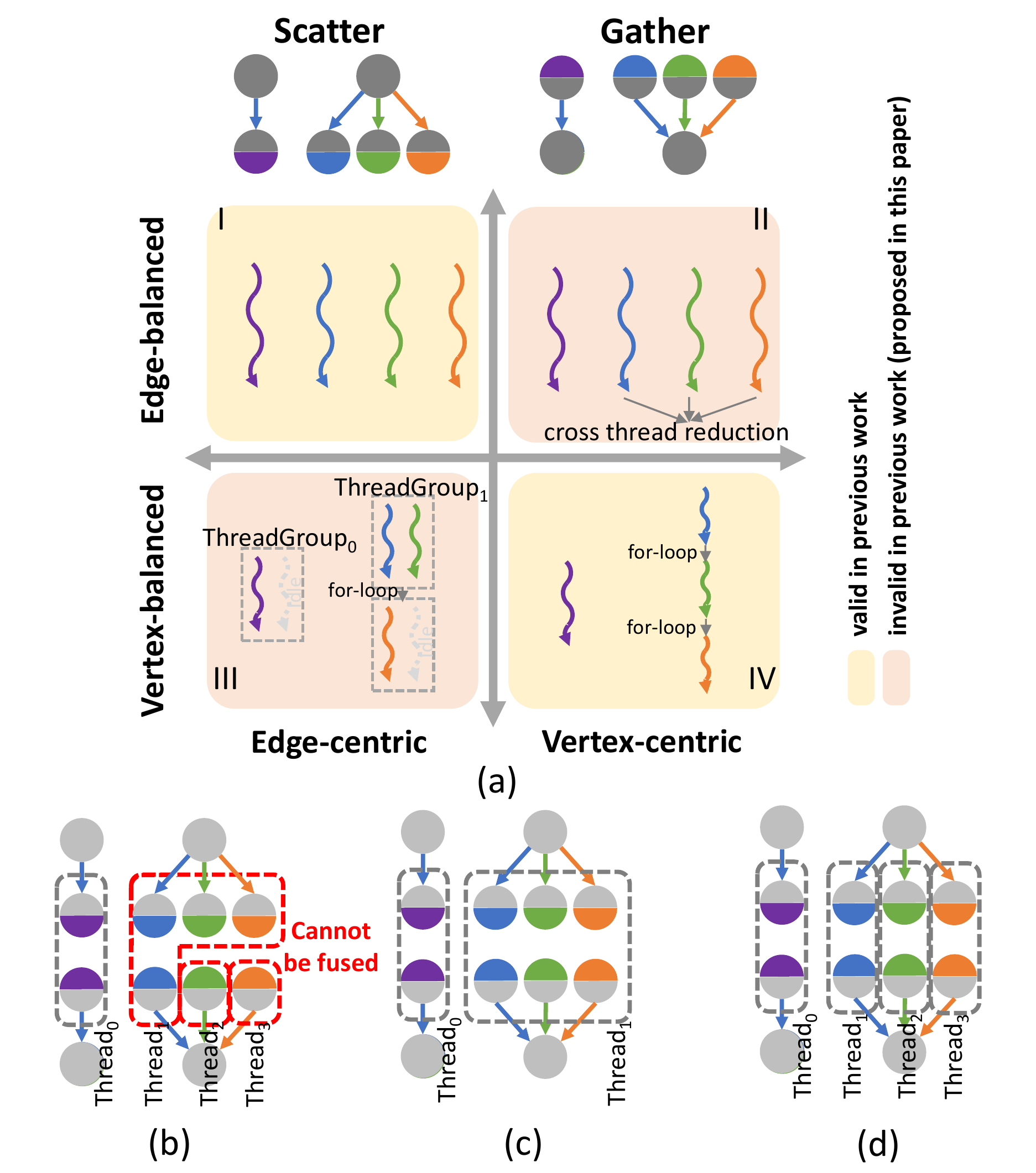}
    \vspace{-25pt}
    \caption{Diagram of the unified thread mapping. (a) We enable different thread mapping schemes for different graph operators. (b) A {\Scatter} with the edge-balanced mapping cannot be fused with a {\Gather} with the vertex-balanced mapping. (c) Vertex-balanced fusion. (d) Edge-balanced fusion.}
    \vspace{-20pt}
    \label{fig:fusion}
\end{figure}

\textbf{Challenge.} The challenge in applying fusion to graph-related operators is the diverged thread-mapping schemes between edge-centric and vertex-centric operators. By edge-centric, we mean the operator whose output is edge features, and by vertex-centric the ones producing vertex features. For example,  {\Scatter} is an edge-centric operator, {\Gather} being vertex-centric, and {\FusedEtoE} and {\FusedVtoV} are hybrid of both. We find current GNN systems commonly implement edge-centric operators in edge-balanced thread-mapping, and vertex-centric ones in vertex-balanced thread mapping. As shown in Figure~\ref{fig:fusion}(a)\uppercase\expandafter{\romannumeral
1}, edge-balanced thread mapping bind parallel workers to different edges. This parallelization strategy naturally matches the edge-centric operator: imagine each worker independently calculate the features for different edges, with no cross-thread communication involved and perfect work-balancing. On the other hand, vertex-balanced thread mapping bind parallel workers to different vertices. This strategy suits the {\Gather} operator because the reduction can be carried by the same worker via a sequential loop as Figure~\ref{fig:fusion}(a)\uppercase\expandafter{\romannumeral
4}. Although the above two strategies are reasonable when seen separately, the issue comes up when we try to fuse operators with different thread-mapping schemes. As shown in Figure~\ref{fig:fusion}(b), the edge-balanced scheme in {\Scatter} and the vertex-balanced scheme in {\Gather} prohibits reusing the intermediate data in the thread's local scope, because the same thread is assigned to an edge at first but a vertex next. 

\textbf{Insight.} Our key insight is that thread-mapping schemes can be decoupled from the operator type: edge-centric operator can also apply vertex-balanced mapping and vise versa. We illustrate these two  scenarios in Figure~\ref{fig:fusion}(a)\uppercase\expandafter{\romannumeral2} and \uppercase\expandafter{\romannumeral3}. To apply vertex-balanced mapping to edge-centric operator, each worker is assigned to loop over the incoming-edge set of a vertex and compute features for these edges. We can increase the number of threads in the same group to exploit parallelism, since features for each edge can be calculated in parallel. The example in Figure~\ref{fig:fusion}(c) reveals a potential issue of imbalanced workload, but the issue is minor as long as we have enough parallelism to fully occupy the GPU, and worth taking if it enables kernel fusion and saves excessive IO.  On the other hand, when applying edge-balanced mapping to vertex-centric operator, we need to handle the cross-thread reduction shown in Figure~\ref{fig:fusion}(d). Cross-thread reduction can be implemented on GPU via atomic arithmetics. Observe that edge-balanced mapping improves workload balancing, but atomic arithmetics can introduce overhead, which we need to compare against the benefit of kernel fusion. 

\textbf{Approach.} Following our insight that both edge-balanced and vertex-balanced schemes can be applied to all operators, we propose to eagerly fuse all graph-related operators with unified thread mapping. By the phrase graph-related, we refer to all operators except the {\ExpensiveApply} ones such as linear projection. In the GAT example, the sequence of {\Scatter}, {\FusedEtoE}, {\FusedVtoV} all fall into this definition, and we are able to fuse them into one single kernel by applying same thread-mapping. In  general, we can select between  vertex-balanced or edge-balanced mapping based on performance profiling. A special case is when {\FusedEtoE} is involved: since an intermediate vertex-feature needs to be reused between two operators, we can only apply the vertex-centric mapping and buffer the vertex-feature in the GPU shared-memory. 


\textbf{Example.} In GAT, there are three graph-related operator that have a potential to fuse: {\Scatter}, {\FusedEtoE} and {\FusedVtoV}. As {\FusedEtoE} requests vertex-centric mapping, we apply unified vertex-balanced mapping to fuse these three operators into one kernel, which saves excessive IO. Assuming one GAT layer has $h$ heads and a feature length of $f$, before operator fusion, the IO of these graph-related operators is $4|\mathcal{E}|h$ for {\Scatter}, $3|\mathcal{E}|h$ for {\FusedEtoE}, and $3|\mathcal{E}|hf+|\mathcal{V}|hf$ for {\FusedVtoV}, with a total of $|\mathcal{V}|hf+7|\mathcal{E}|h+3|\mathcal{E}|hf$. With intermediate data reused, the IO is reduced to $|\mathcal{V}|hf+5|\mathcal{E}|h+2|\mathcal{E}|hf$.

\section{Reducing Memory: Intermediate Data Recomputation for Training} \label{sec:recompute}
\textbf{Motivation.}
GNN systems suffer from excessive memory consumption, because all the intermediate feature tensors are saved for the backward pass. Section.~\ref{sec:fusion} described our techniques to fuse all graph-related operators in the forward pass. Fusion saves not only IO but also memory since no intermediate tensors need to be written-out and read-in. We intend to extend operator fusion for the back-propagation based training scenario to reduce memory consumption. 

\begin{figure}
    \centering
    \includegraphics[width=0.48\textwidth]{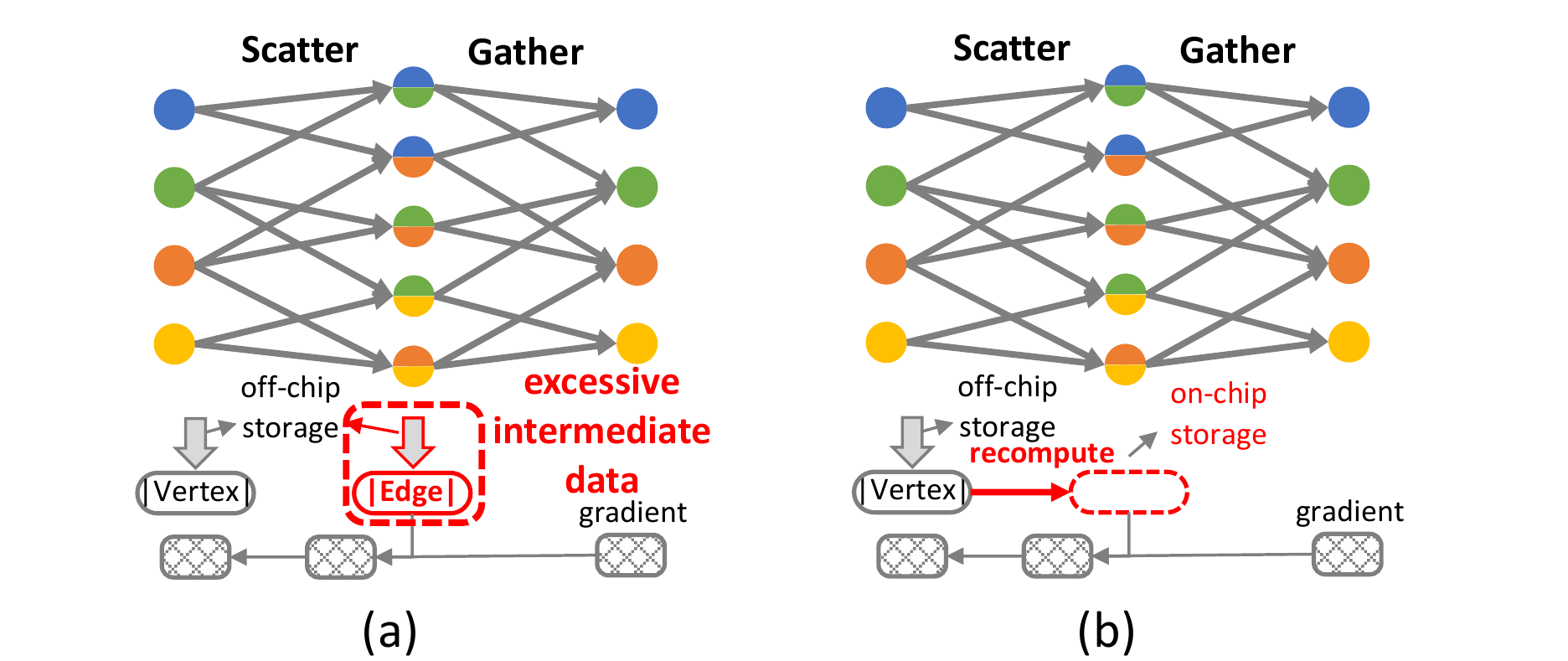}
    \vspace{-25pt}
    \caption{Diagram of the intermediate data recomputation. (a) Edge features are stored for the backward propagation. (b) Edge features are recomputed without storing in the off-chip memory.}
    \vspace{-20pt}
    \label{fig:recompute}
\end{figure}

\textbf{Challenge.} The challenge of avoiding saving intermediate data is back propagation. The role of intermediate data is two folds: (1) it passes the values on the forward computational graph; (2) it passes the intermediate features in the forward pass to the backward computational graph for gradients computing. We can fuse operators both in forward and backward pass, which solves (1). But this is not enough for training, as intermediate data are still needed for backward.

Take Figure \ref{fig:recompute}(a) as an example, which shows a toy example composed of one {\Scatter} step and one {\Gather} step, with operator fusion technique already applied. For the forward pass, we've successfully eliminated the $O(|\mathcal{E}|)$ intermediate data produced by {\Scatter} with operator fusion technique by fusing the {\Scatter}-{\Gather} procedure into one operator, in which the values of the intermediate data are temporarily stored in on-chip memory instead of the off-chip memory. But as we still need this intermediate data for backward propagation, we have to stash the intermediate data in off-chip memory.

\textbf{Insight.} Our key insight is that we can trade memory with computation: if the intermediate data is memory consuming but light weight to compute, we can recompute those needed intermediate data during backward. Based on this, we propose a recomputing technique to deal with the intermediate data in the backward pass, which solves (2). 

\textbf{Approach.} Following our insight that memory can be traded with computation, we propose an empirical criterion $\frac{Computation Cost}{Memory Cost}$ to identify the recomputing opportunity of an operator. If $\frac{Computation Cost}{Memory Cost}$ is no more than $O(1)$, which means we can save one element's memory with no more than one computation, we just recompute the value during the backward pass, because we can save memory with little damage to the runtime latency. Otherwise, we stash the intermediate data as the performance improvement is limited.
In the toy example in figure \ref{fig:recompute}(b), we recompute the $O(|\mathcal{E}|)$ intermediate data instead of stashing it because the computation cost of {\Scatter} is small. By recomputing, we save $O(|\mathcal{E}|)$ memory consumption with $O(|\mathcal{E}|)$ computation. We will show later by experiments that this overhead is usually no more than 10\% in GNN.

\begin{figure*}[!tp]
    \centering
    \includegraphics[width=0.95\textwidth]{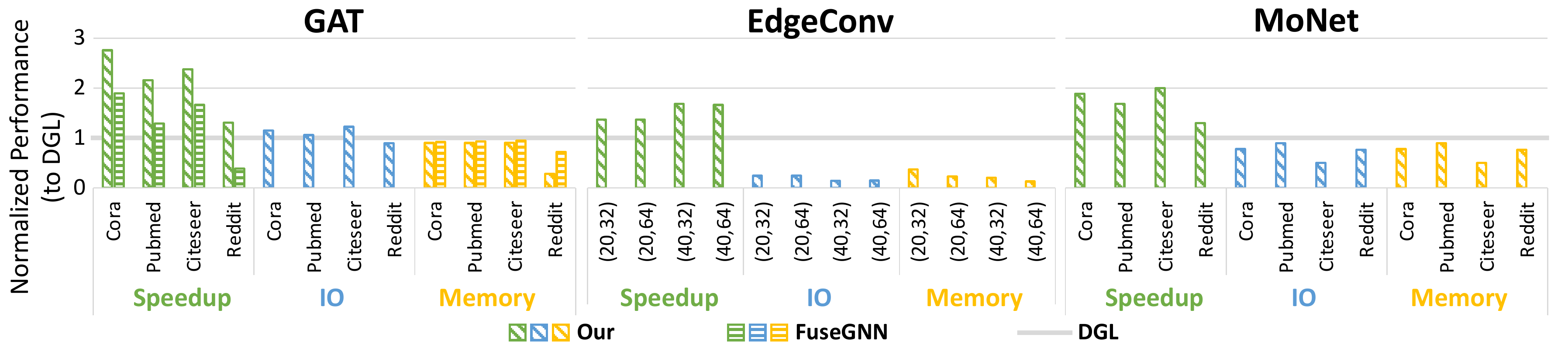}
    \vspace{-7pt}
    \caption{Normalized (to DGL) end-to-end performance on three GNN models from computation (speedup), IO, and memory perspectives.}
    \vspace{-7pt}
    \label{fig:end to end}
\end{figure*}

\textbf{Recomputation with fusion.}
Our recomputing technique usually works for graph-related operators and {\LightApply} operators, which take up much memory space but lightweight to compute. Occasionally, our proposed fusion technique is also applied to graph-related operators and {\LightApply} operators. If we perform fusion without recomputation, we have to stash those needed intermediate data, which still costs a lot of memory space. By fusion-recomputation combo, we eliminate those intermediate data in the whole training process.

\textbf{Example.} 
In GAT, three operators are fused: one {\Scatter}, one {\FusedEtoE} (edge-softmax), and one {\FusedVtoV}. So there are two intermediate data we need to handle: output of {\Scatter} and output of $ReduceScatter$, both of which are $O(|\mathcal{E}|)$. As the $\frac{Computation Cost}{Memory Cost}$ of this {\Scatter} is only $O(1)$, we can just recompute it during backward propagation. The {\FusedEtoE} operator edge-softmax first perform reduction to compute the maximums and the sum of all the exponential as denominator, which is a {\Gather}, followed by a $O(1)$ division to compute the final edge value ({\Scatter} and {\MLPEdge}). The recomputing score $\frac{Computation Cost}{Memory Cost}$ is $O(|log\frac{|\mathcal{E}|}{|\mathcal{V}|}|)$ for {\Gather} and $O(1)$ for {\Scatter} and {\MLPEdge}. According to our standard for recomputing, we store all the maximums and denominators during forward, which only takes $O(|\mathcal{V}|)$, and recompute the other results within $O(1)$ time. 
By our proposed recompute technique, two $O(|\mathcal{E}|)$ intermediate data are eliminated with $O(1)$ overhead in latency.

\section{Experiment}\label{sec:exp}
In this section, we implement our proposed techniques and evaluate them on multiple GNN models and datasets. We (1) demonstrate the overall performance improvements; (2) conduct ablation studies to provide detailed analysis on the benefits brought by each technique; (3) evaluate our implementations on devices with smaller DRAM which couldn't fit in without our optimization.

\subsection{Experimental Setup}
\subsubsection{Benchmarks}


\squishlist
    \item \textbf{Graph Attention Network (GAT)}~\cite{gat} is one of the most classic GNN models, which adopts attention mechanisms to learn the relative weights between connected vertices instead of the identical or pre-determined weights. It first {\Scatter} features to edges and compute attention scores with learnable parameters, then perform {\MLPEdge} followed by {\FusedVtoV}.

    \item \textbf{Edge Convolution (EdgeConv)}~\cite{edgeconv} transforms the point clouds into a k-nearest neighbor graph to represent the topological information, in which points are viewed as vertices and their relative position is modeled as edges. It first {\Scatter} vertex features to edges to compute their relative position, then \texttt{Apply} neural operations on edges and performs {\Gather} to generate vertex embeddings.

    \item \textbf{Mixture Model Network (MoNet)}~\cite{gmm} introduces pseudo-coordinates to determine the relative position among vertices to learn the weight function adaptively. It first performs \texttt{ApplyEdge} to compute gaussian kernel, followed by \texttt{Aggregate}.
\squishend

We choose these models because they represent the trend that GNN models will evolve into more diversity and complexity, from static edge value without gradient \cite{gcn, graphsage} to gradient computation on edge feature \cite{gat,gmm,edgeconv}.

\subsubsection{Baselines}

\squishlist

    \item \textbf{Deep Graph Library (DGL)}~\cite{dgl} is one of the mainstream GNN framework on GPUs, which adapts to existing deep learning software such as PyTorch. It outweighs PyG \cite{pyg} in various GNN models. \cite{fusegnn}

    \item \textbf{FuseGNN}~\cite{fusegnn} is a system for GNN training on GPUs with efficient CUDA kernel implementations and applies operator fusion technique. As fuseGNN does not implement EdgeConv and MoNet, we only compare with it on GAT.
    
\squishend

\subsubsection{Datasets}
For GAT and MoNet, we use four commonly-used GNN datasets for evaluation, including Cora, Citeseer, Pubmed, and Reddit~\cite{gcn,graphsage}. For EdgeConv, we use ModelNet40 classification task with 12,311 meshed CAD models from 40 categories, consisting in predicting the category of a previously unseen shape \cite{modelnet,edgeconv}.

\subsubsection{Platforms \& Metrics}
We implement our proposed technique with a C++ and CUDA backend and a Pytorch-based front-end. Our main evaluation platform is a server with a 10-core 20-thread Intel Xeon Silver 4210 CPU running @ 2.2GHz and an NVIDIA RTX 3090 GPU with CUDA 11. Besides, we use an NVIDIA RTX 2080 GPU to demonstrate our design can achieve comparable performance against RTX 3090.

\subsection{End-to-End Performance}\label{sec:end-to-end}
\textbf{GAT.} As fuseGNN doesn't support multi-head attention, we use the setting: 2 layers with 128 hidden dimensions for evaluation and the end-to-end training results are shown in Figure \ref{fig:end to end}. Compared with DGL, we achieve an average of 2.07$\times$ (up to 2.75$\times$) speedup and save an average of 1.48$\times$ (up to 3.53$\times$) memory consumption. Compared with fuseGNN, we achieve an average of 1.85$\times$ (up to 3.41$\times$) speedup and save an average of 1.29$\times$ (up to 2.55$\times$) less memory consumption. The average IO is increased by 1.3\% due to recomputation. 
On Cora, Citeseer and PubMed, we achieve great speedup mainly because we perform unified vertex-balanced fusion, which is friendly for these datasets. The memory consumption is not greatly saved because what we eliminate is the $O(|\mathcal{E}|)$ intermediate data and the number of edges is small in these datasets. But on Reddit with 233K vertices and 115M edges, we save great memory consumption (3.88GB) compared with DGL (13.7GB) and fuseGNN (9.89GB) mainly because our proposed fusion-recomputation combo eliminates the $O(|\mathcal{E}|)$ intermediate data during training. The memory saving will be more significant if applying multi-head mechanism as in the original paper \cite{gat}.

\textbf{EdgeConv.} We use the same setting as the original paper \cite{edgeconv}: EdgeConv layers=4 with hidden dimensions=\{64, 64, 128, 256\}, the number of nearest neighbors k=20/40, and the batch size=32/64, with a total of four different settings, and the end-to-end training results are shown in Figure \ref{fig:end to end}. Compared with DGL, we achieve an average 1.52$\times$ (up to 1.69$\times$) speedup and save an average of 4.58$\times$ (up to 7.73$\times$) peak memory usage and 5.32$\times$ (up to 6.89$\times$) IO. We apply operator organization and operator fusion technique in EdgeConv. As the {\Gather} function is $\operatorname{max}$, only an $O(|\mathcal{V}|)$ array is needed for back propagation, and recomputation is not applied to further reduce memory consumption. 
Due to the overhead of transforming a point cloud into graph, the end-to-end speedup is not as significant as it should be. However, the memory is largely saved because we optimize the graph-related operators which cause large memory consumption. Note that our memory consumption remains unchanged when k changes, for k is the average number of edges for each vertices. By implementing fusion-recomputation combo, we eliminate all the $O(|\mathcal{E}|)$ intermediate data.

\textbf{MoNet.} We use the setting: 2 layers with 16 hidden dimensions, k=3 r=2 for Cora, k=3 r=3 for Pubmed and Citeseer, k=2 r=1 for Reddit, where k is the gaussian kernel size and r is the dimension for pseudo coordinates in gaussian mixture model. As shown in Figure \ref{fig:end to end}, compared with DGL, we achieve an average of 1.69$\times$ (up to 2.00$\times$) speedup and save an average of 1.47$\times$ (up to 3.93$\times$) peak memory usage and 1.30$\times$ (up to 2.01$\times$) IO. Different from GAT, MoNet doesn't have {\Scatter} in the beginning, therefore operator reorganization is not needed.


\begin{figure}[!tp]
    \centering
    \includegraphics[width=0.48\textwidth]{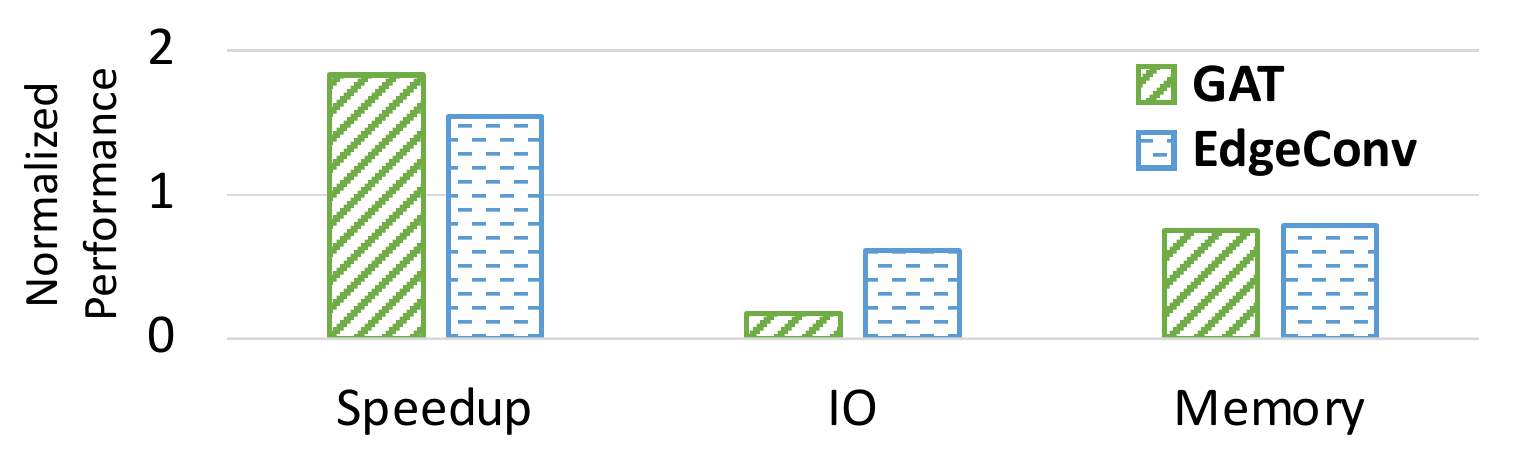}
    \vspace{-25pt}
    \caption{Normalized performance improvements brought by propagation-postponed operator reorganization.}
    \vspace{-20pt}
    \label{fig:reorg-result}
\end{figure}

\subsection{Ablation Studies}\label{sec:ablation}
Without special declaration, we use the setting as follows. (1) GAT: head=4 with feature dimension=64, on Reddit. (2) EdgeConv: k=40, batch size=64, layer=4 with hidden dimensions=\{64, 64, 128, 256\} for training, layer=1 with feature dimensions=64 if only forward. (3) MoNet: k=2, r=1 with feature dimension=16, on Reddit.

\textbf{Reorganization.} Figure \ref{fig:reorg-result} illustrates the benefits of operator reorganization for reducing computation, IO, and memory consumption in GAT and EdgeConv. MoNet has no {\Scatter} and therefore no need for operator reorganization. Due to memory limitation of our device, we evaluate GAT with Pubmed. The baseline is implemented with {\Scatter} before {\MLPEdge}, and the implementation with operator reorganization postpone {\Scatter} and perform {\MLPVertex} first. To clearly show the impacts brought by operator reorganization, We use forward pass for comparison. The experiment results are consistent with theoretical analysis: as redundant computation is eliminated, latency is reduced and redundant IO caused by redundant computation is also eliminated; as we perform {\MLPVertex} before {\Scatter}, one $O(|\mathcal{V}|)$ and one $O(|\mathcal{E}|)$ intermediate data are generated, but if we perform {\Scatter} first followed by {\MLPEdge}, two $O(|\mathcal{E}|)$ intermediate data are generated. For the forward pass, operator reorgnization improves latency by 1.68$\times$, IO by 3.06$\times$, and peak memory usage by 1.30$\times$ on average.

\begin{figure}
    \centering
    \includegraphics[width=0.48\textwidth]{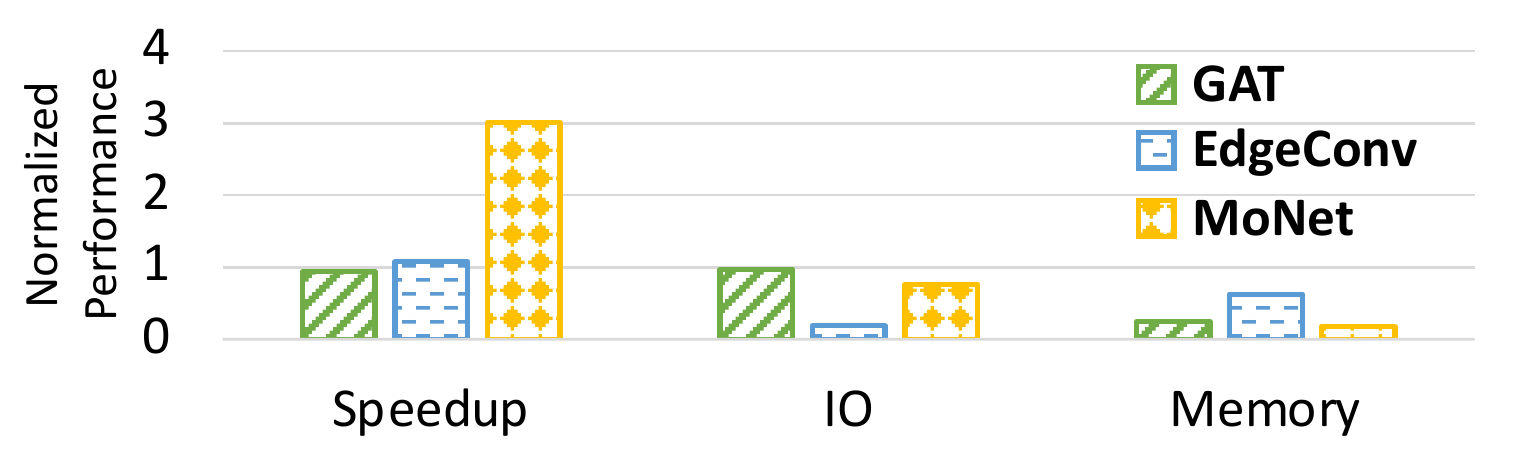}
    \vspace{-25pt}
    \caption{Normalized performance improvements brought by unified thread mapping operator fusion.}
    \label{fig:fusion-result}
    \vspace{-20pt}
\end{figure}

\textbf{Fusion.} Figure \ref{fig:fusion-result} illustrates the benefits brought by operator fusion. We fuse all the graph-related operators with our proposed unified thread mapping scheme, and our proposed fusion technique can be applied to all of these three models. More details about our implementation can be found in appendix. We use For GAT, fusion has a little negative impact on latency, slightly reduces IO and greatly reduces memory consumption. As we use shared memory to perform operator fusion, which introduces extra overhead and Reddit is a very unbalanced graph, the latency is still largely determined by the unbalanced workload after performing fusion. As the neural operators consumes the major part of IO, the relative IO reduction is not significant. The absolute value of IO reduction and memory reduction are about same level. For EdgeConv, IO and memory consumption are greatly reduced, and latency is slightly improved, mainly because of saving write-in and read-out for intermediate data. As the absolute value of IO in EdgeConv is much smaller that GAT, the relative IO reduction is much more significant. For MoNet, latency, IO, and memory are all significantly saved, mainly because of the largely improved data locality and saving for broadcast. For the forward pass, the operator fusion technique improves latency by 1.68$\times$, IO by 1.16$\times$ (up to 5.45$\times$), and peak memory usage by 4.92$\times$ on average.

\textbf{Recomputation.}
Figure \ref{fig:recompute-result} illustrates the benefits brought by intermediate recomputation on GAT and MoNet. As the {\Gather} function in EdgeConv is $\operatorname{max}$, only the indices of the maximum have to be stashed (which is $O(|\mathcal{V}|)$) and there is no need for recomputation. We use three implementations for comparison: (1) without our unified thread mapping operator fusion technique; (2) with the fusion technique but without recomputation technique, which means intermediate data have to be stashed; (3) with both our proposed fusion technique and recomputation technique. For GNN training, only fusion cannot reduce memory consumption, as even if we eliminate some intermediate data during the forward pass with operator fusion, we still need to stash them to perform back propagation. However, with our proposed recomputation technique, we can also eliminate those intermediate data during backward propagation at a small cost of computation. In GAT, recomputation saves 2.21$\times$ memory at the cost of slowing down by 7.1\%. In MoNet, recomputation saves 1.55$\times$ memory and accelerates by 5.9\%.


\subsection{Evaluation on Different GPUs}
\begin{figure}[!tp]
    \centering
    \includegraphics[width=0.48\textwidth]{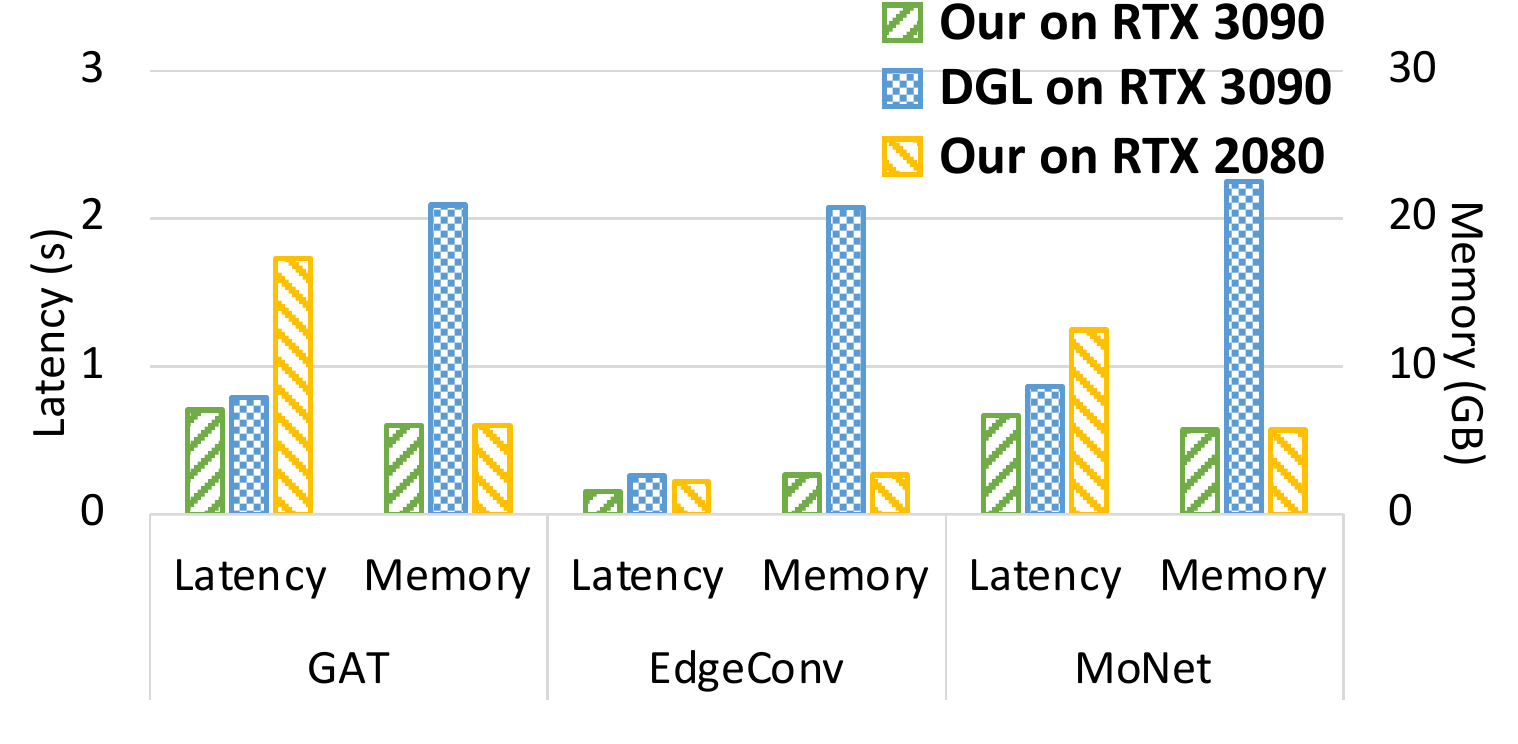}
    \vspace{-25pt}
    \caption{End-to-end performance on different GPUs. Our designs enable running large-scale GNN models with an NVIDIA RTX 2080 GPU, which require the newest NVIDIA RTX 3090 GPU, with a comparable latency.}
    \vspace{-20pt}
    \label{fig:2080}
\end{figure}

With our proposed three techniques, we are able to perform the same training task on devices with much smaller memory capacity. We evaluate our models with the same setting as Section \ref{sec:ablation} on RTX 2080, all of which cannot be done without our proposed techniques due to memory capacity limits. Figure \ref{fig:2080} show that our implementation on RTX 2080 can even achieve 1.17$\times$ end-to-end speedup over DGL on RTX 3090 with 7.73$\times$ less memory for EdgeConv.

\section{Related work}\label{sec:relatework}
\subsection{GNN Systems}
NeuGraph~\cite{neugraph} first introduces SAGA (Scatter, ApplyEdge, Gather and ApplyVertex) abstraction to describe GNNs. It is the first system that bridges the gap between graph processing systems and DNN systems. After that, GNN systems can be categorized as following types: 

\begin{figure}[!tp]
    \centering
    \includegraphics[width=0.48\textwidth]{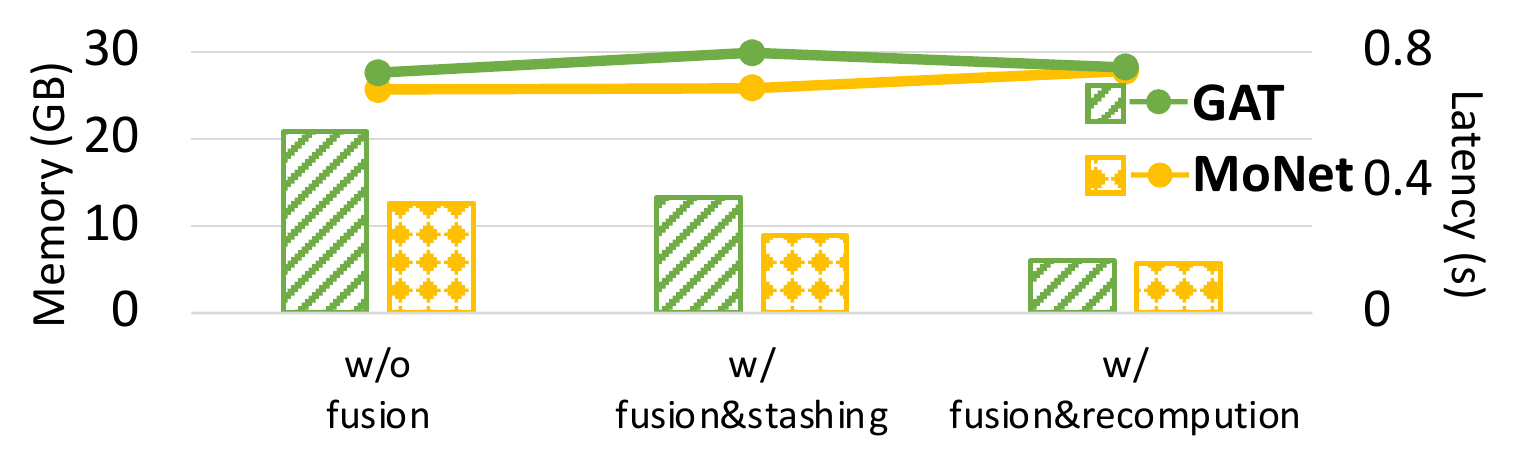}
    \vspace{-25pt}
    \caption{Benefits and overhead brought by intermediate data recomputation. ``w/o fusion": disable fusion. ``fusion\&stashing": fuse operators but stash the needed intermediate data for backward. ``fusion\&recomputation": perform operator fusion as well as recomputation.}
    \label{fig:recompute-result}
    \vspace{-20pt}
\end{figure}

\textbf{GNN computation graph optimization} includes operator reorganization, operator fusion, data flow optimization, etc., and many efforts have been made to solve the challenges in optimizing GNN computation graph: \textbf{(1) Redundant neural operator computation.} Prior work attempts to tackle the computation redundancy via manually modifying the operator combinations to a functionally-equivalent but efficient version. For example, DGL~\cite{dgl} provides a GAT implementation in its GNN-module library, where the {\MLPEdge} (the linear projection) is separated into two functions applied to vertex-features ahead of propagation. However, a theory inside this practice needs to be extracted for optimizing similar scenarios, as we do in this paper. \textbf{(2) Inconsistent thread mapping.} Fusion is widely used in conventional Deep Neural Networks (DNNs)~\cite{dnnfusion}. FuseGNN~\cite{fusegnn} manages to fuse any two edge-centric operators, but lacks the technique to fuse a vertex-centric operator with an edge-centric one, which we address in this paper via unified thread mapping. \textbf{(3) Excessive intermediate data.} Huang \textit{et al.,}~\cite{ppopp2021} reduces intermediate data during forward but cannot handle back propagation because the intermediate data are missed. FuseGNN~\cite{fusegnn} stashes the intermediate data during forward, but lacks the recomputation technique, which still consumes great memory space.

\textbf{GNN runtime optimization} includes neighbor grouping, graph reordering etc, which introduces a preprocessing procedure to schedule the workload assignment and memory layout. GNNAdvisor \cite{gnnadvisor} and Huang \textit{et al.,}~\cite{ppopp2021} both utilize neighbor grouping to balance the workloads among GPU threads and blocks and exploit memory locality. GNNAdvisor further use Rabbit Reordering \cite{rabbit} to maximize the graph modularity by clustering. By neighbor grouping and graph reordering, the runtime workload balance and memory locality are improved by introducing some preprocessing overhead.


\subsection{DNN Systems}
TASO \cite{taso} proposes a novel computation graph optimizer for DNNs that can automatically generate graph substitutions. DNNFusion \cite{dnnfusion} proposes a set of fusion methodologies to work in conjunction with computation graph rewriting for DNN inference. Chen \textit{et al.,}~\cite{chen2016training} introduces the recomputation technique to DNN training to trade computation with memory. Our proposed operator reorganization technique is more of eliminating computing redundancy, while DNN computation graph substitution is more of finding a better substitution. Our unified thread mapping operator fusion technique is also different from operator fusion in DNNs, as GNN introduces \pleasecheck{graph-related} operator, which brings about the divergent thread mapping between edge-centric and vertex-centric operators. And unlike DNN recomputation, which incurs roughly 30\% of additional latency \cite{chen2016training}, overhead by our proposed recomputation technique is 
$<$10\% as we utilize the characteristics of GNN training.

\section{Conclusion}\label{sec:conclusion}
In this paper, we present a thorough study of GNN computational graph optimization. We point out GNN systems suffer from redundant neural operator computation, inconsistent thread mapping, and excessive intermediate data. We propose a systematic framework with propagation-postponed operator reorganization, unified thread mapping for fusion, and intermediate data recomputation. We achieve up to 2.75$\times$ end-to-end speedup, 6.89$\times$ less memory IO, and 7.73$\times$ less memory consumption over state-of-the-art frameworks. We even enable running large-scale GNN models with an NVIDIA RTX 2080 GPU, which require the newest NVIDIA RTX 3090 GPU, with a comparable latency. More specifically, we provide an optimization-friendly perspective to understand GNN computational graph, which can be extended to other hardware platforms.

\bibliographystyle{mlsys2021}
\clearpage
\section*{Appendix}
\subsection*{A\hspace{5mm} GNN Operators} \label{sec:appendix:op}

This section formally describes our taxonomy of GNN operators, briefly introduced in Section 2.1 as 4 basic operators: {\Scatter}, {\Gather}, {\MLPEdge}, {\MLPVertex}, and 2 high-level operators: {\FusedEtoE} and {\FusedVtoV}. We further illustrate how to construct popular GNN models from this set of operators.

\subsubsection*{A.1\hspace{5mm} Operator Definition} 

Let a graph be $\Gchar=(\Vchar, \Echar)$, where $\Vchar$ represents the set of vertices, and $\Echar$ represents the set of edges. The elements in $\Echar$ is tuples of $(u, e, v)$, where $u,v \in \Vchar$ and $e$ is a unique id. The tuple $(u,e,v)$ indicates there is an edge indexed by $e$ pointing from $u$ to $v$. \footnote{Here we assume directed edges, but can generalize the theory to undirected edges by seeing each edge $u\leftrightarrow v$ as two directed ones $u\rightarrow v$ and $v \rightarrow u$.} We define four basic operators as follows: 

\textbf{\Scatter}: $m_e = \phi (h_u, h_v), (u,e,v) \in \Echar$.
For every edge, perform a binary operation (function $\phi(\cdot, \cdot)$) on the features attached to the two vertices that the edge connects to.

\textbf{\Gather}: $h_v = \psi (\{m_e: (u,e,v) \in \Echar \})$.
For every vertex, perform a reduction operation to the features attached to all edges that connects to it. 

\textbf{\MLPEdge}: $m^{new}_e = f_e (m_e[, m'_e, \cdots]), (u,e,v) \in \Echar$. For every edge, perform the same function $f_e$ that transforms its current feature (and any history features). This operator is \graphirr, meaning that its outcome does not change if the graph structure (connections) changes. 

\textbf{\MLPVertex}: $h^{new}_v = f_v (h_v[, h'_v, \cdots]), v \in \Vchar$. For every vertex, perform the same function $f_v$ that transforms its current feature (and any history features). This operator is also \graphirr like {\MLPVertex}.

Through composing the above four operators, we also propose two high-level operators that are widely seen in GNN models:

\textbf{\FusedVtoV}: \\
$h^{new}_v = \psi(\{f_e(\phi(h_u, h_v), m_e)\}), (u,e,v) \in \Echar$. 
It is a sequence of three basic operators: {\Scatter} to generate edge features, {\MLPEdge} to transform the edge feature or combine it with any history features, and finally {\Gather} to reduce edge features and generate new vertex features. A typical example is the neighborhood feature-reduction in vanilla GCN, where each vertex takes the sum of all its neighbor-vertices' features, essentially $h^{new}_v = sum (\{w_e \cdot h_u: (u, e, v) \in \Echar\})$. This step can be expressed by {\FusedVtoV} by binding $\phi$ as copying source-vertex's feature, $f_e$ as multiplying the edge weight $w_e$, and $\psi$ as summation.

\textbf{\FusedEtoE}:\\
$m^{new}_e = f_e(\phi(\psi(\{m_{e}\}),h_u), m'_e), (u,e,v) \in \Echar$.
It is a sequence of three basic operators: {\Gather} to reduce edge features into vertex features based on the vertex's adjacent edge group, and {\Scatter} to broadcast the reduction results to all edges, and finally {\MLPEdge} to combine the broadcast values and any history features into new edge features. This operation can be used when the edge features are normalized within a neighborhood set, as happens in the edge-softmax. Edge-softmax performs $m^{new}_e = softmax(\{m_e': (u\in \mathcal{N}(v), e', v\} )[e]$, where 
$$softmax(x_1, \cdots, x_n)[i] = \frac{e^{(x_i - \max_{k}(x_k))}}{\sum^n_{j=1}{e^{(x_i - \max_{k}(x_k))}}}$$. This step can be expressed by the following code snippet: 

\texttt{RS}1: $\psi \leftarrow max$, $\phi \leftarrow \text{copy}$, $f_e \leftarrow \text{substraction}$,\\
\texttt{RS}2: $\psi \leftarrow sum$, $\phi \leftarrow \text{copy}$, $f_e \leftarrow \text{division}$.

\subsubsection*{A.2\hspace{5mm} Construct GNN Models}

\textbf{GCN}

\begin{figure*}
    \centering
    \includegraphics[width=0.98\textwidth]{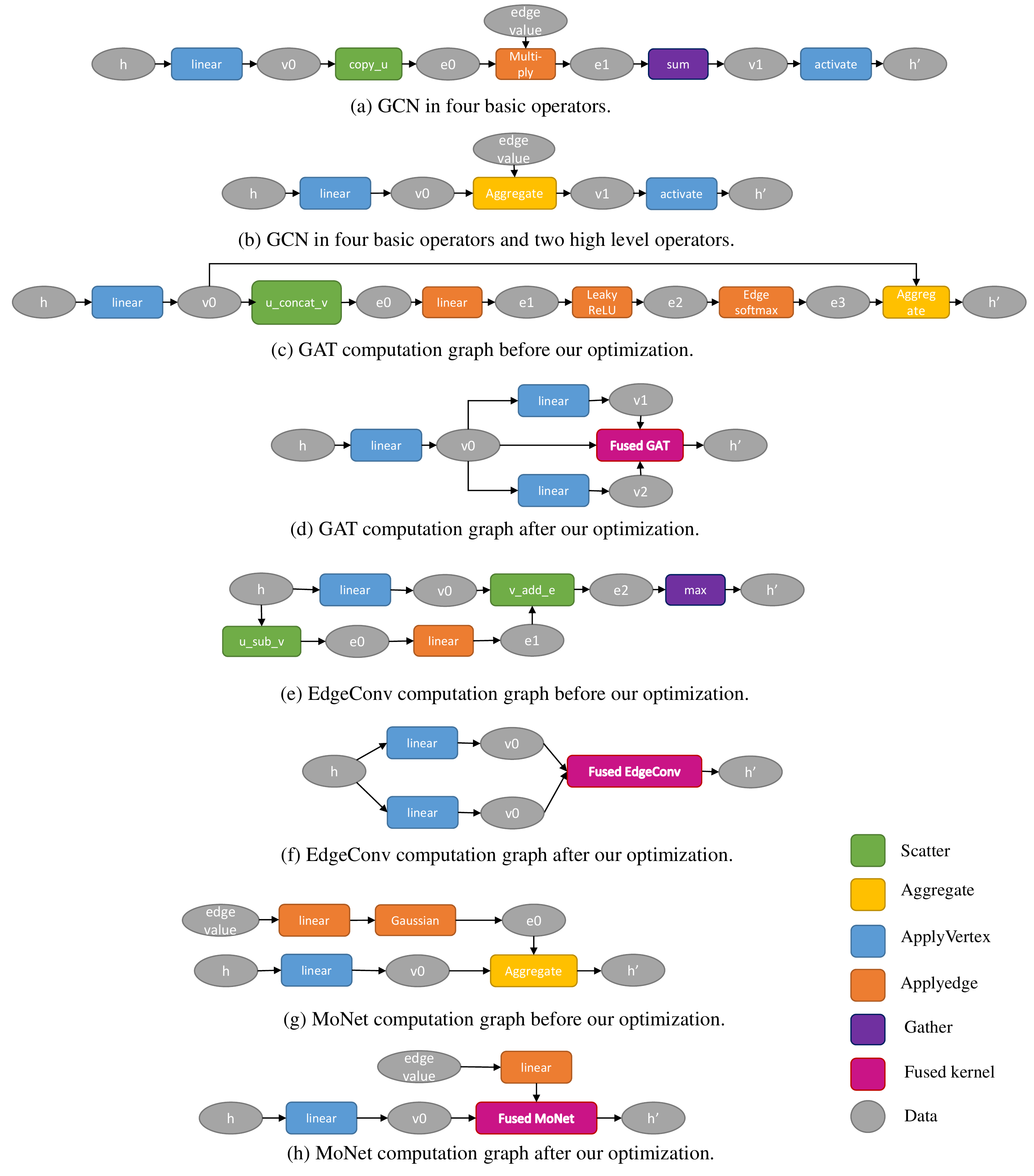}
    \caption{Construct GNN models with GNN operators.}
    \label{fig:appendix_graph}
\end{figure*}

Vanilla GCN is defined as:
$$
h_{v}^{(l+1)}=\sigma\left(b^{(l)}+\sum_{u \in \mathcal{N}(v)} {e_{u v}} h_{u}^{(l)} W^{(l)}\right)
$$
where $\sigma$ is an activation function, $b$ is a bias, and $W$ is weight to learn. With four basic operators, we first perform {\MLPVertex}, then copy source vertex's feature to edges ({\Scatter}) and multiply the edge weights ({\MLPEdge}) to obtain $e_{u v} h_{u}^{(l)} W^{(l)}$, followed by a gather with summation ({\Gather}) and an activation (\MLPVertex), as shown in figure \ref{fig:appendix_graph}(a). Figure \ref{fig:appendix_graph}(b) shows how to describe the same procedure with an high-level opeartor \FusedVtoV.

\textbf{GAT}

GAT is defined as:
$$
    h_{v}^{(l+1)}=\sum_{u \in \mathcal{N}(v)} e_{uv} W^{(l)} h_{u}^{(l)}
$$
$$
    e_{i j}^{l} =\operatorname{edge-softmax}\left(\operatorname{LeakyReLU}\left(\vec{a}^{T}\left[W h_{i} \| W h_{j}\right]\right)\right)
$$
where $W$ and $a$ are learnable parameters. Figure \ref{fig:appendix_graph}(c) shows one way to compute this. 
Assume the input node feature vectors are concatenated into a feature matrix $\boldsymbol{H}^{(l)} \in \mathbb{R}^{n \times f^{(l)}}$, and operator reorganization technique is already applied. We first perform a dense matrix matrix multiplication to transform this feature matrix into $\widetilde{\boldsymbol{H}^{(l)}}=\boldsymbol{H}^{(l)}\times \boldsymbol{W}^{(l)} \in \mathbb{R}^{n\times f^{(l+1)}}$ with torch.nn.linear. We decompose the weight vector ${\boldsymbol{a}}\in \mathbb{R}^{2f^{(l+1)}}$ into $\left[\boldsymbol{a}_l||\boldsymbol{a}_r \right]$ and compute attention scores $\boldsymbol{A}_l=\widetilde{\boldsymbol{H}^{(l)}}\times\boldsymbol{a}_l\in \mathbb{R}^{n \times 1}$ and $\boldsymbol{A}_r=\widetilde{\boldsymbol{H}^{(l)}}\times\boldsymbol{a}_r\in \mathbb{R}^{n \times 1}$.

After that, $\boldsymbol{M}_0\in\mathbb{R}^{n} $ are generated by $$\boldsymbol{M}_0=\operatorname{u\_add\_v}(\boldsymbol{A}_l,\boldsymbol{A}_r)$$
An \MLPEdge operator is then applied to generate $$\boldsymbol{M_1}=\operatorname{LeakyReLU}(\boldsymbol{M}_0) \in\mathbb{R}^{n} $$ followed by a \texttt{ReduceScatter} operator to generate $$\boldsymbol{M_2}=\operatorname{edge\_softmax}(\boldsymbol{M}_1)\in\mathbb{R}^{n} $$
An \FusedVtoV operator is performed to generate $$\boldsymbol{H}^{(l+1)}=\operatorname{reduce\_sum}(\boldsymbol{M_2},\widetilde{\boldsymbol{H}^{(l)}}) \in \mathbb{R}^{n \times f^{(l+1)}}$$
In our implementation, we fuse the computation of $\boldsymbol{M}_0,\boldsymbol{M}_1,\boldsymbol{M}_2,\boldsymbol{H}^{(l+1)}$ into one operator, as shown in figure \ref{fig:appendix_graph}(d).

\textbf{EdgeConv}

Figure \ref{fig:appendix_graph}(e) shows one way to compute EdgeConv. The mathematical definition of one EdgeConv layer is
$$
    h_{v}^{(l+1)}=\max _{u \in \mathcal{N}(v)}\left(\Theta \cdot\left(h_{u}^{(l)}-h_{v}^{(l)}\right)+\Phi \cdot h_{v}^{(l)}\right)
$$
where $\mathcal{N}(v)$ is the neighbor of $v . \Theta$ and $\Phi$ are linear layers. In SOTA gnn framework DGL, one edgeconv layer is computed as shown in figure \ref{fig:appendix_graph}(e). Define the input node feature matrices as $ \boldsymbol{H}^{(l)}\in\mathbb{R}^{n\times f}$. The $(h_u^{(l)}-h_v^{(l)})$ is computed by

$$\boldsymbol{E}^{(l)}=\operatorname{u\_sub\_v}\left(\boldsymbol{H}^{(l)}\right)\in\mathbb{R}^{e\times f^{(l)}}$$

followed by one linear {\MLPEdge}
$$\boldsymbol{E}_\Theta^{(l)}=\Theta \cdot \boldsymbol{E}^{(l)}\in\mathbb{R}^{e\times f^{(l+1)}}$$

An linear {\MLPVertex} is performed to compute $\Phi \cdot h_{v}^{(l)}$:
$$\boldsymbol{N}_\Phi^{(l)}=\Phi \cdot \boldsymbol{H}^{(l)}\in\mathbb{R}^{n\times f^{(l+1)}}$$

followed by

$$\boldsymbol{E}_{\Theta+\Phi}^{(l)}=\operatorname{e\_add\_v}\left(\boldsymbol{E}_\Theta^{(l)},\boldsymbol{N}_\Phi^{(l)}\right)\in\mathbb{R}^{e\times f^{(l+1)}}$$

In the end, a reduce function is called to update the node features

$$\boldsymbol{H}^{(l+1)}=\operatorname{reduce\_max}\left(\boldsymbol{E}_{\Theta+\Phi}^{(l)}\right)\in\mathbb{R}^{n\times f^{(l+1)}}$$







\textbf{GMMConv}

GMMConv is defined as:
$$
m_{uv}=f(x_u,x_v),x_u \in \mathcal{N}(v)
$$
$$
w_k(m)=exp(-\frac{1}{2}(m-\mu_k)^T\Sigma_k^{-1}(m-\mu_k))
$$

$f$ here is a linear projection, $\Sigma_k$ is a covariance matrix of the gaussian kernel, $\mu_k$ is the mean of the gaussian kernel. By setting covariance matrix and mean as parameters with gradient, GMMConv could learn weight $w_k$ in training process (\texttt{ApplyEdge}). 
$$
h_v^{(l+1)}=\frac{1}{K}\sum_{u \in \mathcal{N}(v)} \sum_k^K w_k(m_{uv}){h_u}_k^{(l)}
$$
To get node feature, GMMConv multiplies node embedding with gaussian weight, followed by gathering the sum of multi-kernels of embeddings (\texttt{Gather}).

\subsection*{B \hspace{5mm} Back-propagation of GNN Operators}\label{sec:appendix:backprop}

\begin{figure*}[htbp]
    \centering
    \includegraphics[width=0.99\textwidth]{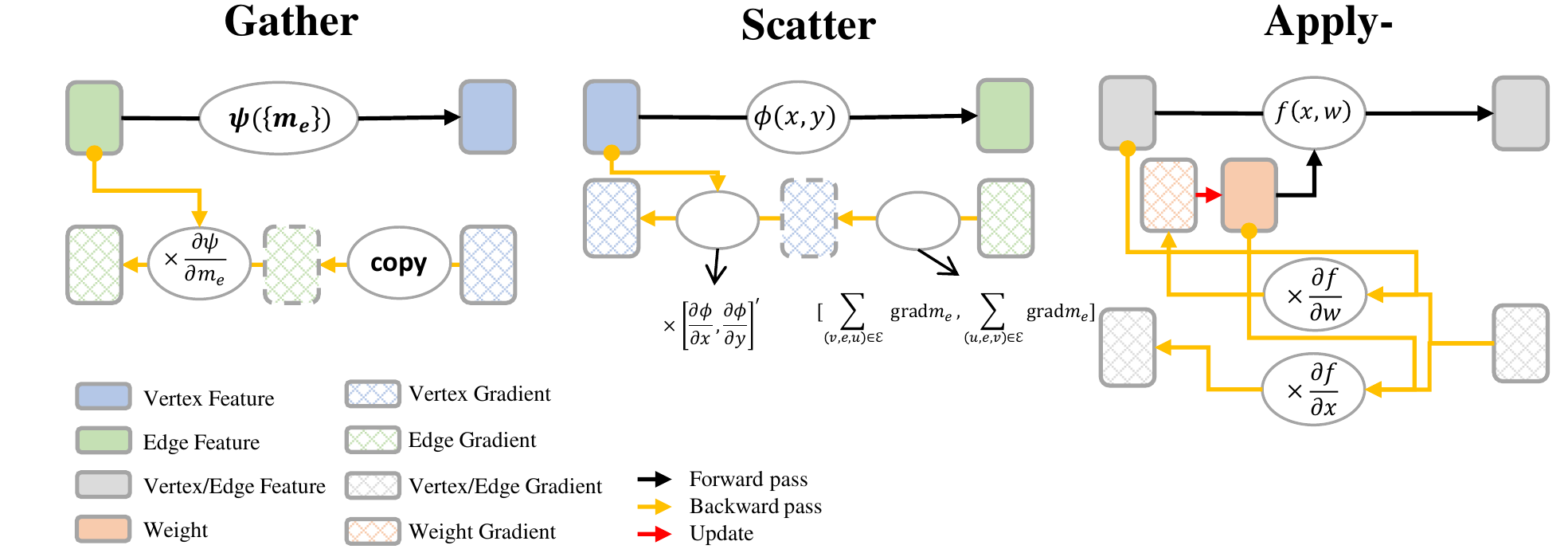}
    \caption{Back-propagation dataflow of GNN operators.}
    \label{fig:apdx_backprop}
\end{figure*}

In this subsection, we derive the backward pass of the four GNN operators, and show that they can still be constructed by the four basic operators. Here $\circ$ represents composition of operators where the latter operator gets applied first.

\textbf{\Gather}: The backward pass of {\Gather} is a {\Scatter} followed by an {\MLPEdge}. 
{
\begin{equation*}
    \begin{aligned}
    \text{\textbf{Forward:  }} h_v =& \psi (\{m_e: (u,e,v) \in \Echar \}), \\
    \text{\textbf{Backward:  }} \grad{m_e} =& \grad{h_v}\times \frac{\partial \psi}{\partial m_e}\\
    =&\MLPEdge_{f_e \leftarrow ({\times \grad{m_e}})} \\
    &\circ \Scatter_{\phi \leftarrow \text{copy\_v}}
    \end{aligned}
\end{equation*}
}

\textbf{\Scatter}: The backward pass of {\Scatter} is a {\Gather} followed by an {\MLPVertex}. 
{
\begin{equation*}
    \begin{aligned}
    \text{\textbf{Forward:  }} m_e =& \phi (h_u, h_v), (u,e,v) \in \Echar, \\
    \text{\textbf{Backward:  }} \grad{h_v} =& \sum_{(v,e,u)\in \Echar}\grad{m_e}\times \frac{\partial \phi}{\partial h_v}\\
    & +  \sum_{(v,e',u')\in \Echar}\grad{m_{e'}} \times \frac{\partial \phi}{\partial h_u} \\
    =&\MLPEdge_{f_e \leftarrow (\times [\frac{\partial \phi}{\partial h_v}, \frac{\partial \phi}{\partial h_u}]^T)}\\
    & \circ \Gather_{\phi \leftarrow [\sum\grad{m_e}, \sum\grad{m_{e'}}]}
    \end{aligned}
\end{equation*}
}

\textbf{\MLPs}: The backward pass of \graphirr {\MLPs} is also \graphirr, and can be derived in the same way as operators in neural networks. 
{\begin{equation*}
    \begin{aligned}
        \text{\textbf{Forward:   }} y =& f(x, w) \\
        \text{\textbf{Backward:   }} \grad{w} =& \grad{y}\times \frac{\partial f}{\partial w} \\
        &\grad{x} = \grad{y}\times \frac{\partial f}{\partial x}
    \end{aligned}
\end{equation*}
}
Hence the backward of {\MLPs} is two {\MLPs}, one calculating the gradient of input and one for the gradient of weight parameters.

Figure~\ref{fig:apdx_backprop} visualizes the forward-backward dataflow of each GNN operator.


\end{document}